# An AI-Powered Research Assistant in the Lab:
# A Practical Guide for Text Analysis Through Iterative Collaboration with LLMs

Gino Carmona-Díaz[1,2,3], William Jiménez-Leal[2,3], María Alejandra Grisales[1], Chandra Sripada[4,5,6], Santiago Amaya[3,7], Michael Inzlicht[8,9] & Juan Pablo Bermúdez[1,10,*]

[1] Social and Human Sciences Faculty, Universidad Externado de Colombia, Bogotá, Colombia
[2] Psychology Department, Universidad de los Andes, Bogotá, Colombia
[3] Laboratorio de Emociones y Juicios Morales, Universidad de los Andes, Bogotá, Colombia
[4] Department of Psychiatry, University of Michigan, Ann Arbor, MI, USA
[5] Department of Philosophy, University of Michigan, Ann Arbor, MI, USA
[6] Weinberg Institute for Cognitive Science, University of Michigan, Ann Arbor, MI, USA
[7] Department of Philosophy, Rice University, Houston, TX, USA
[8] Department of Psychology, University of Toronto, Toronto, Canada
[9] Rotman School of Management, University of Toronto, Toronto, Canada
[10] Department of Philosophy, University of Southampton, Southampton, UK

[*] Corresponding author. Email: j.bermudez@soton.ac.uk

## Abstract

Analyzing texts such as open-ended responses, headlines, or social media posts is a time- and labor-intensive process highly susceptible to bias. However, LLMs are promising tools for text analysis, using either a predefined (top-down) or a data-driven (bottom-up) taxonomy, without sacrificing quality. Here, we present a step-by-step tutorial to efficiently develop, test, and apply taxonomies for analyzing unstructured data through an iterative and collaborative process between researchers and an LLM. Using personal goals provided by participants as an example, we demonstrate how to write prompts to review datasets and generate a taxonomy of life domains, evaluate and refine the taxonomy through prompt and direct modifications, test the taxonomy and assess intercoder agreements, and apply the taxonomy to categorize an entire dataset with high intercoder reliability. We discuss the possibilities and limitations of using LLMs for text analysis.

**Keywords:** Text analysis, Qualitative Analysis, Taxonomies, Large Language Models, Artificial Intelligence, GPT, Tutorial.

1. **Introduction**

Large Language Models (LLMs) are transformer neural networks trained on internet-scale data that exhibit sophisticated abilities, including human-like linguistic fluency. LLMs are increasingly used by social scientists to automate diverse tasks in fields such as medicine (Thirunavukarasu et al., 2023), law (Surden, 2023; Schwarcz & Choi, 2023), and education (Jeon & Lee, 2023).

Psychology is no exception. LLMs are increasingly applied in various tasks within psychological practice and research (Demszky et al., 2023). In terms of clinical applications (Carlbring et al., 2023), LLMs have been used to estimate suicide risk in patients (Lee et al., 2024; Levkovich & Elyoseph, 2023), to generate empathic responses that outperform human experts (Ovsyannikova et al., 2025), and to identify and treat cases of depression (AlSamhori et al., 2024) or neurodevelopmental disorders (Bertacchini et al., 2023). In terms of research applications, LLMs have been used to translate surveys (Kunst & Bierwiaczonek, 2023), generate items to assess cognitive abilities (Sayin & Gierl, 2024) or personality traits (Götz et al., 2023; Lee et al., 2023; Hernandez & Nie, 2023), and score tasks to evaluate abilities such as creativity (DiStefano et al., 2024) or divergent thinking (Dumas et al., 2024). Some studies have even explored the use of LLMs in *conducting* psychological research, such as proposing hypotheses (Banker et al., 2024; Hermida-Carrillo et al., 2024), executing experiments with simulated agents (Park et al., 2023), writing papers (Lockwood & Castleberry, 2024), and evaluating abstracts as peer reviewers (Shcherbiak et al., 2024).

One of the applications of LLMs in psychological and social science research is automated text analysis for the purposes of qualitative research (see a review, see Smirnov, 2024). Although unstructured data, such as social media posts, news headlines, or open-ended responses, can provide valuable insights for social scientists, analyzing such data can be challenging due to the significant time and effort required to train coders to code, score, categorize, and classify the data (Iliev et al., 2015). Manual analysis, such as grounded theory coding or thematic analysis, requires extensive close reading and is limited by human cognitive constraints. The labor-intensive nature of this work typically forces researchers to limit either the number of participants or the number of responses collected, constraining the scale and scope of qualitative investigations. Additionally, researchers engaged in these tasks may also display biases, prejudices, or fatigue, which can compromise the quality of results (Campbell et al., 2013; Hruschka et al., 2004).

At the same time, simple computer-automated methods like word frequency analysis and dictionary-based approaches, which simplify qualitative analysis by counting the frequencies of specific terms, have limitations that may impact the analysis. E.g., their dependency on a specific dataset of words may lead to the incorrect identification of key concepts due to contextual nuances, sarcasm, or homonyms (false positives). These automated analyses are also known to have problems recognizing conceptual allusions without explicit keywords (false negatives) (Cook & Jensen, 2019; Iliev et al., 2015; Eichstaedt et al., 2021). LLMs have the potential to overcome these challenges and facilitate more efficient and accurate qualitative data analysis.

Several recent studies have explored the performance of LLMs in categorizing unstructured data, including open-ended responses (Prinzing et al., 2024; Mellon et al., 2024;

Santana-Monagas et al., 2024), tweets, news headlines (Rathje et al., 2024; Gilardi et al., 2023), vignettes (Levkovich & Elyoseph, 2023), and academic papers (Patra et al., 2023). These studies have revealed promising results, with LLMs outperforming supervised learning methods (machine learning techniques trained on labeled data to learn the relationship between inputs and outputs) and achieving accuracy levels comparable to those of human coders.

More significantly, LLMs are promising tools for developing taxonomies to classify unstructured information (Moraes et al., in prep). By *taxonomy* we mean a structured and methodical system for classifying complex phenomena along one or more conceptual dimensions. Researchers commonly develop taxonomies to classify different observations into a set of categories that shed light on their nature, causes, and implications, thus reducing the complexity of phenomena and facilitating analysis to better understand them (Bradley et al., 2007). Even though researchers often define taxonomies *ex ante* based on a set of theoretical considerations, conceptual dimensions, and purpose of the study (i.e., using a top-down approach), data-driven approaches in which categories are formed in a bottom-up manner based on the data to guide text analysis might be a preferable approach when theoretical considerations are too abstract or subject to controversies, when conceptual dimensions are poorly understood, or when the purposes of study are compatible with multiple alternative categorization schemes (Eichstaedt et al., 2021). For example, researchers studying personal goals might begin with an established top-down taxonomy—such as the intrinsic versus extrinsic goal distinction—but find that many real-world goals don't fit cleanly into those categories. A bottom-up approach, informed by participants' own language and priorities, can supplement and refine the initial framework to better reflect the structure of the data and the social reality it captures.

Compared to manual and NLP methods, a collaborative and iterative LLM approach offers several advantages (Table 1). Analyzing text through an iterative collaboration with LLMs enables the processing of large amounts of information with an optimal level of contextual understanding, consistency, and flexibility, while requiring minimal human effort and time. Although it may not be the indicated option in all cases (for example, if a researcher aims to develop an original theory based on data analysis, grounded theory may be more appropriate), it is nonetheless a valuable alternative for many research tasks.

To support this growing interest and showcase the possibility of automating text analysis for social research, we propose a step-by-step tutorial for developing, testing, and applying taxonomies. In order to automate taxonomy construction (partially or fully) with the help of LLMs, it is necessary to break down taxonomy construction into clear steps with appropriate instructions for each. To this end, we provide an eight-step guide (see Figure 1), accompanied by materials, to formulate, test, and use a taxonomy to analyze open-ended responses. This process combines the iterative and collaborative efforts of human researchers and LLMs. Although the tutorial is designed specifically for GPT-4.0, the steps and recommendations can be applied to any of the widely known LLMs. Since each step's instructions are designed to be readable by both humans and LLMs, they could be applied even to the design of taxonomies without AI tools. To illustrate the tutorial's steps, we use as an example a dataset of personal goals provided by participants in a series of studies on goal pursuit and socioeconomic status carried out in Colombia during 2024–2025.

Table 1. Comparison between approaches for text analysis

| Feature | Manual Grounded Theory/Thematic Analysis | Traditional NLP Methods | LLM-Assisted Approach |
| --- | --- | --- | --- |
| Scale capability | Limited (dozens to hundreds of units) | High (thousands to millions) | High (thousands to millions) |
| Contextual understanding | High (human comprehension) | Low-Medium (depends on algorithm) | High (contextual models) |
| Resource requirements | High human effort, low computational | Low human effort, medium computational | Medium human effort, high computational |
| Development time | Weeks to months | Days to weeks (with expertise) | Days (with proper prompting) |
| Classification consistency | Variable (dependent on coder) | High (algorithmic) | High (with proper testing) |
| Flexibility | High (easily adjusted by humans) | Low (requires reprogramming) | High (adjusted through prompts) |
| Theory development | Strong (primary purpose) | Weak (primarily descriptive) | Medium (requires human interpretation) |

First, we describe the open-ended response data that will be used to illustrate the steps of the tutorial. Then, we detail the eight steps of taxonomy building, offering definitions, warnings, recommendations, examples, and materials. We then reflect on the scope and limitations of LLM-generated taxonomies. Finally, we conclude by discussing future possibilities, potential challenges, and ethical considerations of using LLMs in research. All materials mentioned in this tutorial can be accessed here: https://osf.io/ahwsv/

## 2. Research example context

To illustrate the process of developing, testing, and applying a taxonomy to analyze text with the assistance of LLMs, we use data from our studies on personal goals. In these studies, we sought to examine the relationship between socioeconomic status (SES) and personal goal achievement. Specifically, we aim to identify what factors explain why individuals from lower socioeconomic backgrounds experience greater difficulties accomplishing their goals. Previous findings in the literature suggest that lower-SES individuals are more likely to engage in counterproductive behaviors that undermine their chances of personal goal achievement (Shurtleff 2009; Banerjee & Duflo 2007; Skiba & Tobacman 2008; Blalock et al. 2007; Haisley et al. 2008; World Health Organization2011). Several theories have explained this tendency by referencing a culture of poverty (Lewis, 1966), deficient human capital (Lusardi & Mitchell, 2014), tunnel thinking focused on scarce goods (Mullainathan & Shafir, 2013; Mani et al., 2013), the effects of stress and negative affect (Haushofer & Fehr, 2014), low levels of aspirations (Dalton et al., 2016), and a rational response to a current or anticipated lack of financial resources (Hilbert et al., 2022). However, we argued for a perspective that emphasizes the epistemic effects of poverty, where adverse conditions of scarcity themselves serve as evidence that justifies engaging in future-discounting behaviors.

To test the different possible explanations, we conducted four studies —two cross-sectional and two longitudinal— in which we asked 1.759 participants to report one or more personal goals they expected to achieve in the coming months, along with additional information about them including questions regarding participants' perceptions of ambition, self-efficacy, response efficacy, and locus of control about the goals, the plans designed to achieve them, and open-ended question regarding general information about the goals. Additionally, in the longitudinal studies, we asked participants about different conflicts they experienced as they pursued their goals as well as their perceived progress. In total, we collected more than 3.185 goals from individuals across a range of socioeconomic levels. We aimed to explore the relationship between SES and different goal-related variables such as progress, self-efficacy, and locus of control.

One factor that may explain or mediate differences in goal achievement across socioeconomic status (SES) groups is the type of goals individuals pursue. If individuals from lower SES backgrounds disproportionately pursue widely different goal types, we might attribute differences in goal attainment to differences in goal type. Although a well-known taxonomy of personal aspirations and goals exists (Kasser & Ryan, 1993, 1996, 2001; Grouzet et al., 2005), this taxonomy was developed top-down by researchers and may not be applicable to the study's dataset. For example, it lacks a category for personal aspirations related to education and gaining knowledge—a recurrent goal in our study—but includes a category on hedonism that is heavily focused on sexual pleasure— a type of goal that no participant in our study reported. In response to these limitations, we aimed to develop a taxonomy of personal goals that reflects our study particpants' actual aspirations, is constructed through a bottom-up approach, and meets the specific requirements of our research objectives.

However, the large number of personal goals collected posed a significant challenge. Limited temporal and cognitive resources, combined with the risk of overlooking important nuances, could affect the taxonomy's structure and accuracy. Although we could have used a random subset of goals to develop the taxonomy, doing so risked omitting critical examples necessary to define relevant categories. Considering these challenges, we explored the use of LLMs to support the development of a taxonomy of personal goals capable of identifying the types of goals pursued by individuals across different socioeconomic groups.

### 3. Step-by-step tutorial

Our eight-step taxonomy development, test and application procedure is illustrated in the flowchart in Figure 1. The procedure includes: (1) creating the initial taxonomy-generation prompt; (2) generating the taxonomy; (3) evaluating it; (4) editing the prompt (if necessary); (5) adjusting the taxonomy; (6) testing the taxonomy; (7) assessing disagreements and adjusting (if necessary); and (8) applying the rubric to the dataset. Each of the eight steps can be followed by human or LLM-based coders and can be used with any LLM. We will present examples of these steps from our research on personal goals.

Figure 1. *Flowchart for developing, test and using taxonomies with LLMs or humans*

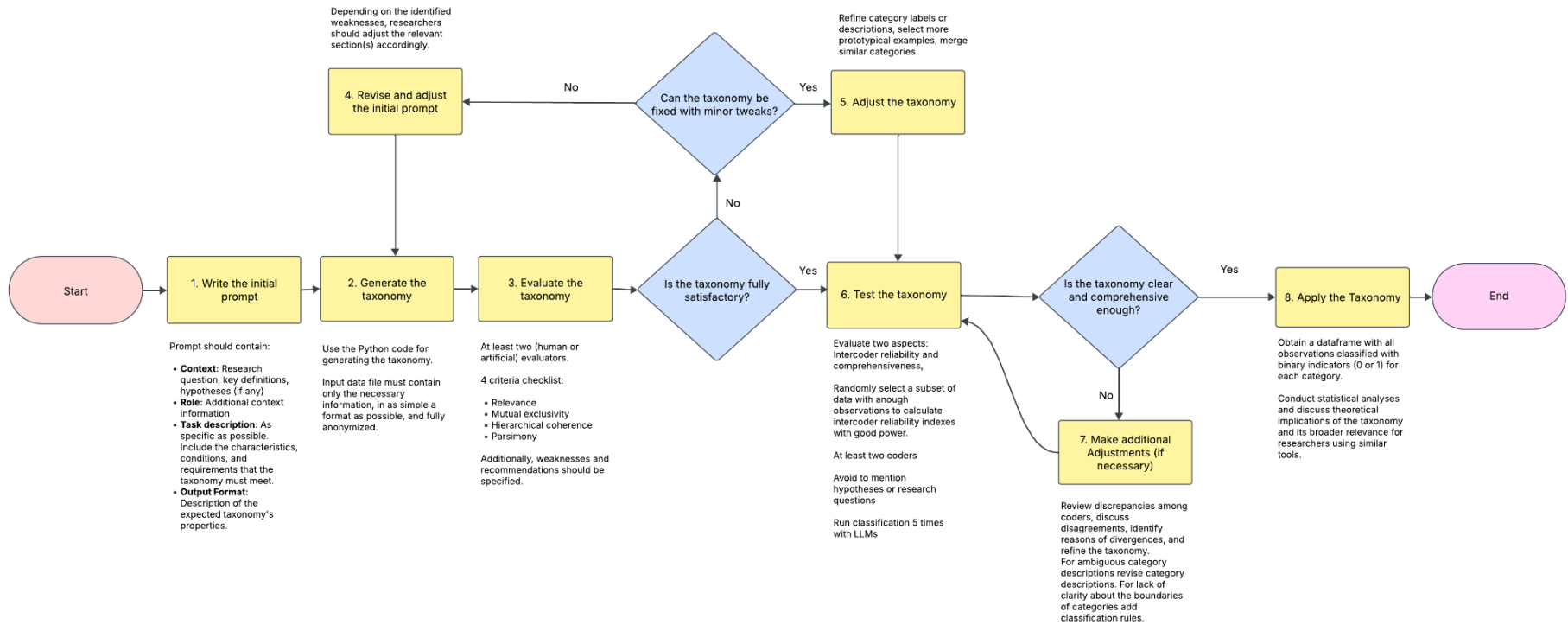

**Step 1. Write an initial prompt**

Regardless of whether the taxonomy will be developed by an artificial or a human agent, the first step is to write a clear and complete prompt that clearly states the key information relevant to building the taxonomy. The initial prompt must be clear, specific and concise while including four key aspects: context, role, task, and expected output.

*1.1. Context*

First, researchers must specify the *context* of the request, including information about the research question the taxonomy is meant to help answer, any hypotheses to be tested through the classification (if any have been identified), and the definitions of concepts necessary to develop a taxonomy. Countless taxonomies can be formulated based on the same dataset, so contextual information about the specific research project is crucial to develop a useful taxonomy that avoids the inclusion of irrelevant categories.

It is also important to mention details about the characteristics of the data that will serve to develop the taxonomy: the sources of information, to which question or stimuli the responders answered, and what kind of responses were expected. In studies with open-ended questions, irrelevant, short and absurd responses are not uncommon. Therefore, general information about the data to analyse is key to helping coders identify which responses are rich, relevant inputs and which are irrelevant and may be safely ignored.

*1.2. Role*

Second, researchers must specify the *role* the coder will perform. In this case, it can be described as an assistant of a team of social researchers, collaborating on the development of categories to analyze qualitative data drawn from interviews, open-ended responses, or field notes. Providing role information gives coders (particularly LLMs) additional contextual information about the request and the expected output, improving the quality and relevance of the resulting taxonomy.

*1.3. Task*

Third, researchers must clearly define the *task* the LLMs are expected to carry out. The task description should be as specific as possible, as an ambiguous command is unlikely to generate responses that satisfy the researchers' needs. This description should include the characteristics, conditions, and requirements that the taxonomy must meet. For example, it should specify whether the taxonomy should be hierarchical (with different levels of abstraction and some categories subsumed under others) or flat (with all categories with the same level of abstraction and no overlap), whether overlapping between categories are allowed, how exhaustive the taxonomy should be, and the desired number of categories (if there is one). Researchers may also provide additional elements, such as step-by-step instructions or examples of a well-constructed taxonomy.

1.4. *Expected output*

Finally, the prompt should specify the *expected output*, both in terms of format and content. Since the resulting taxonomy will be included in subsequent prompts for LLM-based or human coders, it is advisable to request that categories be something akin to a bullet point list, each with an identifying label, a brief but precise definition, and one or more examples taken directly from the dataset. This will allow researchers to understand and evaluate taxonomy. However, the expected output may take a different format and include different information depending on the researcher's interests.

Since developing these instructions is ultimately an exercise in prompt engineering, it is worth keeping in mind that prompt engineering is a new and rapidly developing field[1]. The study of interactions with AI is only just beginning, so advances and improvements in best practices are likely to continue emerging. Further best practices can be found in the official prompt engineering guide by [OpenAI](OpenAI) or [Claude](Claude). Box 1 contains a sample prompt, which we used to generate our taxonomies for the personal goals dataset, and the supplemental materials include a prompt template.

Box 1. Initial prompt

---

**Context**

We are studying the relationship between socioeconomic status and the achievement of personal goals. We define personal goals as the outcomes a person aims to achieve, guided by their needs, interests, values, and aspirations.

Previous studies have shown that people with fewer economic resources face greater difficulties in making progress toward their goals. Our objective is to explore which variables might explain why this happens.

To address this, we conducted a longitudinal study in which participants from different socioeconomic backgrounds described their personal goals. Some time later, we evaluated the progress they had made toward these goals.

One possible explanation for the differences in progress may be related to the types of goals each group pursues. Therefore, we need to classify the reported goals to identify the kinds of goals people in our study are pursuing.

Each participant provided:

- A basic description of their goal (what they want to achieve).
- Additional information (why the goal is important, how long they have been pursuing it, and how they feel about it).

---

[1] A search in Scopus with the "prompt engineering" term revealed that 27 works were published in 2022, 346 in 2023, and 1.376 in 2024.

> **Role**
>
> You are an AI assistant helping an expert in social psychology who is studying the relationship between socioeconomic status and the pursuit of personal goals.
>
> **Task description**
>
> Review all the goals provided by the participants and build a taxonomy that classifies the personal goals according to the life domains to which the goals belong.
>
> The taxonomy must meet the following criteria:
>
> - Clarity and distinction: Create clear and well-differentiated categories. Avoid redundancies or categories that are too similar.
> - Grouping of equivalents: If you identify goals that are essentially the same (for example, "Improve physical health" and "Take care of my physical condition"), group them under a single category.
> - Relevance: Do not include irrelevant categories. If you find meaningless or nonsensical responses, assign them to a category called "Not applicable" instead of creating a new category.
> - Conceptual independence: Do not rely on existing taxonomies as references. Build the classification solely based on the information provided in the participants' goals.
>
> **Output format:**
>
> Present each category of the taxonomy as a bullet point, and for each category include:
>
> - A clear and descriptive name.
> - A brief definition of the category.
> - Two concrete examples taken directly from the provided goals.

**Step 2. Generate the Taxonomy**

Once the prompt has been written, it can be given to coders—human or artificial—along with the dataset to develop the taxonomy. In terms of providing information to an LLM-based coder, this tutorial is tailored to GPT-4.0, though researchers can also test it with other models. Given the potentially large size of the data, ChatGPT may reject the request, so researchers must obtain an API key from the OpenAI platform and submit the request through Python. The supplementary materials include a commented [Python code in Colab](#) that can be copied and adapted to different research goals.

The initial prompt must be provided along with an input data file that contains only the necessary information in as simple a format as possible and fully anonymized. Although LLMs can process large and broad datasets, they remain fallible, and complex input requests may increase the likelihood of hallucinations and other errors (Zhang et al., 2024; Smirnov, 2024). Additionally, as we argue in the Discussion, LLMs are not free of biases. Therefore, including irrelevant or unnecessary information in the input can lead to biased outputs. In our example, we provided GPT with a data frame containing two columns: one with participant IDs and another with the full text used for category creation—basic goal descriptions and additional goal information (see the dataframe in the Supplemental Material). For example:

> Basic description: 'Lose 3 kilograms of weight'. Additional information: 'It is important because although I am aware that I have a healthy weight, my weight has normally been my current weight minus 3 kg. At my age, I wouldn't want this increase to cause disadvantages in managing my health. I have been wanting to achieve this for 5 months. I feel that I am capable of reaching it.'

For analyses requiring the processing of more complex information chains, a strategy to simplify the input without losing the richness of the context of the information is to create a single character string that integrates all pieces of data into a coherent account. For instance, in a different analysis, we developed a taxonomy to classify self-control strategies. We utilized data including participant IDs, personal goals, conflicts encountered in pursuing those goals, and the strategies employed to manage them. Rather than supplying a four-column dataset, we crafted narratives that integrated all relevant information into single character strings:

> The person R_3CyFBsUrVKge2BS has the goal of "Passing an exam". Recently, they reported that when they had an opportunity to move toward this goal, they faced a temptation conflict. At that moment, they needed "To study for the exam" but "they felt like watching TV". To overcome this, they implemented the following strategy: 'Moving to another room to avoid getting distracted by the TV.

After processing the 3,185 personal goals in our dataset, GPT proposed an 11-category taxonomy. Box 2 shows the resulting taxonomy obtained in this step, along with a definition and a couple of examples per each category.

Box 2. Resulting taxonomy.

---

**Language Learning:** Goals related to acquiring proficiency in a new language for personal, professional, or travel purposes (e.g., Learn English to travel to another country and interact without problems; Learn Portuguese to maintain cognitive skills and for personal and professional development).

**Education and Learning:** Goals related to completing educational programs, acquiring new knowledge, or learning new skills (e.g., Finish my university studies because it is the career I chose and have always liked; Learn English to communicate my products to my audience and for job opportunities).

**Financial Stability and Independence:** Goals focused on achieving financial security, independence, or improving economic conditions (e.g., Achieve financial stability to live more freely without worrying about money; Get out of debt to improve my economy and gain financial freedom).

**Career and Professional Development:** Goals related to advancing in one's career, acquiring new skills, or achieving professional recognition (e.g., Change jobs to experience new challenges and grow professionally; Graduate as a professional to have greater opportunities in the job market).

---

> **Entrepreneurship and Business:** Goals related to starting, growing, or managing a business for personal or financial reasons (e.g., Start my own business to provide more peace and well-being to my family; Consolidate my company as a personal achievement).
>
> **Health and Well-being:** Goals aimed at improving physical health, mental health, or overall well-being (e.g., Lose weight for health reasons and to feel strong for myself and my daughters; Exercise more to improve health and physical appearance).
>
> **Family and Relationships:** Goals centered around improving family life, relationships, or supporting family members (e.g., Improve relationship with my son to maintain harmony and trust; Spend more quality time with my partner).
>
> **Travel and Exploration:** Goals involving traveling to new places for leisure, cultural experiences, or personal fulfillment (e.g., Travel to Japan to learn about a new culture and see the world beyond my area; Travel around the world to achieve a higher status in my professional life).
>
> **Housing and Living Environment:** Goals focused on acquiring, improving, or changing one's living situation (e.g., Buy a house to ensure stability and provide a home for my family; Move to a new apartment to reduce expenses and save for a planned trip).
>
> **Personal Development and Fulfillment:** Goals aimed at personal growth, self-improvement, or achieving a sense of fulfillment (e.g., To be a better person every day to feel happy and improve daily; Stop procrastinating to engage in activities that help me grow personally and professionally).
>
> **Not Applicable:** Responses that are meaningless, nonsensical, or do not fit into any relevant category (e.g., Develop telepathy through meditation and practice; None, I don't have to give opinions about my goals).

**Step 3. Evaluate the Taxonomy**

Once the coders have proposed a taxonomy, agents (human and LLM-based) must play the role of evaluators to assess whether the taxonomy is good enough for research purposes. To facilitate the evaluation process, we propose a taxonomy assessment rubric, consisting of a checklist and a qualitative component. Box 3 presents the prompt used to ask GPT to evaluate the taxonomy, which can also be used as the instructions for human researchers in charge of this task. In the checklist component, evaluators assess the taxonomy based on the following criteria:

- **Relevance:** The taxonomy effectively aids in understanding the phenomenon of interest by classifying items in a way that is useful for answering the project's research questions. Its criteria speak to the hypotheses to be tested (if any have been mentioned).
- **Mutual Exclusivity:** The proposed categories at the same level are conceptually distinct, capturing different dimensions or aspects of the data without significant overlap between them.
- **Hierarchical Coherence:** If nested categories are not used, all categories are at the same level of abstraction, meaning that none can reasonably be interpreted as a subcategory of another. On the other hand, if nested categories are used, there is a logical organization across levels, with subcategories representing proper subsets of their parent categories, and each child category fully belonging within its parent's conceptual boundaries.
- **Parsimony:** The taxonomy does not include too many categories. If categories are deleted, important nuance or question-relevant information is likely to be lost. The taxonomy avoids nested subcategories where a simpler flat structure is likely to suffice.

These four criteria seek to form a logically coherent and largely sufficient framework for evaluating taxonomies at the pre-classification stage because they address complementary dimensions of taxonomic quality. Relevance ensures the taxonomy serves its intended purpose. Mutual Exclusivity and Hierarchical Coherence together establish the logical structure necessary for consistent classification, with the former ensuring horizontal distinction between categories and the latter ensuring vertical integrity across levels. Parsimony acts as a regulating principle, preventing excessive complexity that would undermine both conceptual clarity and practical utility. These criteria aim at balancing theoretical concerns (logical coherence, conceptual distinctiveness) with practical considerations (fitness for purpose, usability) while focusing exclusively on the structural properties of the taxonomy itself—precisely what can and should be evaluated before application to actual data.

Taxonomies can be evaluated using a checklist format, providing a 1 if the taxonomy meets the criteria or a zero if not, along with a brief justification of the response. Additionally, evaluators can conduct a qualitative assessment by identifying the most notable weaknesses of the taxonomy and providing recommendations on how to address them. Box 3 shows a prompt to request the evaluators (human or LLM-based) to evaluate the resulting taxonomy.

Box 3. Instructions for taxonomy evaluation

---

**Context**

We are studying the relationship between socioeconomic status and the achievement of personal goals. We define personal goals as the outcomes a person aims to achieve, guided by their needs, interests, values, and aspirations.

Previous studies have shown that people with fewer economic resources face greater difficulties in making progress toward their goals. Our objective is to explore which variables might explain why this happens.

To address this, we conducted a longitudinal study in which participants from different socioeconomic backgrounds described their personal goals. Some time later, we evaluated the progress they had made toward these goals.

One possible explanation for the differences in progress may be related to the types of goals each group pursues. Therefore, we need to classify the reported goals to identify the kinds of goals people in our study are pursuing.

---

Each participant provided:

- A basic description of their goal (what they want to achieve).
- Additional information (why the goal is important, how long they have been pursuing it, and how they feel about it).

Based on a dataset with all those personal goals, a research assistant developed a taxonomy.

**Role**

You are a researcher in social psychology who is contributing to a project studying the relationship between socioeconomic status and the pursuit of personal goals.

**Task Description**

Review the attached taxonomy and evaluate whether it meets the following criteria:

- Relevance: The taxonomy effectively contributes to understanding the participants' personal goals by classifying observations.
- Mutual Exclusivity: The proposed categories at the same hierarchical level are conceptually distinct, capturing different dimensions or aspects of the data without significant overlap.
- Hierarchical Coherence: All categories should be at the same level of abstraction, meaning none should reasonably be interpreted as a subcategory of another.
- Parsimony: The taxonomy does include a sufficient number of categories so that if any categories were removed, important nuances or information relevant to the research questions would likely be lost, but avoids unnecessary subcategories when a simpler flat structure would suffice.

In addition to these ratings, provide comments on the following:

- Most notable weaknesses of the taxonomy.
- Recommendations for addressing these weaknesses.

**Output Format**

For each criterion, provide a binary response (1 for 'Meets the criterion', 0 for 'Does not meet the criterion'), along with a brief justification of your response. Additionally, provide separate responses for the two qualitative points:

- Weaknesses: What are the most important limitations that this taxonomy has and that prevent it from satisfying the evaluation criteria?
- Recommendations: What are the crucial modifications that could be made to the taxonomy to allow it to better satisfy the evaluation criteria? The recommendations should be as few and as simple as possible.

We recommend that three independent evaluators — both human and artificial agents — review the taxonomy and provide their qualifications and comments. Then, the main researcher must review the evaluations and decide whether the taxonomy is fully satisfactory and requires no additional adjustments and can be tested (Step 6), whether it can be fixed with minor tweaks (Step 5), or whether it is necessary to rewrite the initial prompt due to it presents several limitations (Step 4).

In the Supplemental Material are the evaluations conducted by three evaluators, including both research assistants and LLMs. The three evaluators agreed that the taxonomy fulfilled the Relevance criterion, and two agreed that it fulfilled the Parsimony criterion. However, all evaluators believed that the taxonomy failed to meet the Mutual Exclusivity and Hierarchical Coherence criteria. Specifically, the taxonomy distinguishes between categories that might be merged (e.g., Language Learning and Studying, or Career and Professional Development and Entrepreneurship and Business), and includes overly restricted categories that could be more comprehensive (e.g., Housing and Living Environment).

In addition, one evaluator proposed a new category: Community and Social Impact. While we recognize that people may pursue personal goals related to this and other categories, we aimed to conduct this process bottom-up, based on the goals reported by participants in our sample. Therefore, we opted not to include new categories unless we found several goals that could not be classified under any existing taxonomy category. Taken all evaluations together, we opted to skip to step 5 and just introduce minor changes to test the taxonomy.

**Step 4. Revise and Adjust the Initial Prompt (if Necessary)**

If the taxonomy does not meet all the evaluation criteria and researchers identify significant weaknesses, they should revise the original taxonomy-generation prompt developed in Step 1. Depending on the identified weaknesses, researchers should adjust the relevant section(s) accordingly. Shortcomings in Relevance can often be addressed by adding information to the prompt's Context section. Clarifying the research goals and the taxonomy's purpose can help to ensure the taxonomy effectively addresses the research question.

Limitations related to Mutual Exclusivity, Hierarchical Coherence and Parsimony should be resolved by explicitly specifying what additional requirements the taxonomy must meet the task description. To do this, additional criteria can be included (e.g., limiting the maximum number of categories, stating that categories referring to a particular constellation of concepts like learning languages, finishing a degree, and learning to code) should be kept within a single category instead of splitting throughout multiple categories.

Other weaknesses may necessitate modifications to the role description by specifying particular aspects the coder should consider For example, if the researcher is developing a taxonomy to create a set of options for multiple-choice questions and needs the categories to be clear enough for anyone, the role may be better described as that of a layperson—someone without experience in social research—rather than a research assistant.

Furthermore, researchers can modify any aspect of the input data frame if they identify issues affecting taxonomy generation. For example, if the input is too large—making it difficult for the LLM to capture all the information or leading to hallucinations—researchers may consider simplifying the input by removing unnecessary information or switching from a tabular to a narrative format. Once the initial prompt is revised, researchers can proceed to Step 2 again.

**Step 5. Adjust the Taxonomy**

Even if the evaluation results suggest that modifying the initial prompt is unnecessary, researchers may still introduce some final adjustments to refine the taxonomy itself. The evaluation process may highlight minor shortcomings that, once addressed, lead to an improved version of the taxonomy. For example, minor adjustments could include refining category labels or descriptions, selecting more prototypical examples, or merging two highly similar categories whose distinction does not provide valuable information.

Box 4 presents the adjusted version of the taxonomy after addressing the weaknesses identified in step 3. Specifically, we merged two pairs of categories. Language Learning was subsumed under the Education and Learning category to encompass all goals related to acquiring knowledge and skills, regardless of the subject. Similarly, the Career Development category was broadened to also include the Entrepreneurship and Business category, encompassing all forms of career progression, whether dependent or self-employed. Finally, the Housing and Living Environment category was changed to Material Acquisition to broaden its scope to other types of acquisitions. The final taxonomy includes nine categories.

Box 4. Adjusted version of taxonomy.

> **Education and Learning:** Goals related to completing educational programs, acquiring new knowledge, abilities, or learning new skills (e.g., Finish my university studies because it is the career I chose and have always liked; Learn English to communicate my products to my audience and for job opportunities; learn how to drive).
>
> **Financial Stability and Independence:** Goals focused on achieving financial security, independence, or improving economic conditions (e.g., Achieve financial stability to live more freely without worrying about money; Get out of debt to improve my economy and gain financial freedom).
>
> **Career and Professional Development:** Goals related goals related to professional and work-life development. This would include job promotions, career changes, transitioning out of unemployment, and starting independent work through personal businesses or entrepreneurial ventures (e.g., Change jobs to experience new challenges and grow professionally;start my own business to provide more peace and well-being to my family).
>
> **Health and Well-being:** Goals aimed at improving physical health, mental health, or overall well-being (e.g., Lose weight for health reasons and to feel strong for myself and my daughters; Exercise more to improve health and physical appearance).
>
> **Family and Relationships:** Goals centered around improving family life, relationships, or supporting family members (e.g., Improve relationship with my son to maintain harmony and trust; Spend more quality time with my partner).
>
> **Travel and Exploration:** Goals involving traveling to new places for leisure, cultural experiences, or

> personal fulfillment (e.g., Travel to Japan to learn about a new culture and see the world beyond my area; Travel around the world to achieve a higher status in my professional life).
>
> **Material Acquisition:** Goals aimed at acquiring, retaining, or maintaining material goods. These may include purchasing a first or second home, vehicles, household appliances, leisure or personal enjoyment items, as well as home repairs or improvements (e.g.,Buy my own house; Buy a new car; Finish purchasing the furniture for my apartment.).
>
> **Personal Development and Fulfillment:** Goals aimed at personal growth, self-improvement, or achieving a sense of fulfillment (e.g., To be a better person every day to feel happy and improve daily; Stop procrastinating to engage in activities that help me grow personally and professionally).
>
> **Not Applicable:** Responses that are meaningless, nonsensical, or do not fit into any relevant category (e.g., Develop telepathy through meditation and practice; None, I don't have to give opinions about my goals).

**Step 6. Test the Taxonomy**

After the final adjustments, researchers must test the taxonomy to determine whether it appropriately classifies items from the dataset. Testing is an intermediate step to check whether the taxonomy is clear enough that any coder classifies items correctly.

Since testing the entire dataset may be too time- or resource-consuming, researchers can test the taxonomy using a randomly selected subset of data. Determining the appropriate number of items to analyze is an important consideration. Although there is no clear standard for the ideal test subset size, the decision may be based on the required sample size for analyses involving intercoder reliability indexes and the number of categories.

To assess reliability among three coders, indices may require a minimum sample size of 30 items (Koo & Li, 2016). However, the required sample size may be higher if there are only two coders. Nevertheless, it is always advisable to calculate the statistical power of the metrics using the given sample size and increase the sample if the desired statistical power is not reached (Sim & Wright, 2005), although some indices, such as Krippendorff, are flexible and can be generalized to different numbers of coders and items.

On the other hand, if the number of categories is high, as in our case, at least 10 items per category should be included to ensure sufficient data for classification testing. This number is admittedly arbitrary, and random item selection may lead to some categories having few or no items while others contain many. However, once the subset is tested, researchers may analyze additional data to enhance the classification of underrepresented categories.

It is essential to have multiple coders classify the subset using the taxonomy to test intercoder reliability and ensure that the taxonomy is interpreted correctly and without ambiguities. At this stage, it is also necessary to write a new prompt document. Like the initial prompt, it should include context, role, task description, and expected output. However, since the coders' task in this step involves classifying items rather than developing a taxonomy, certain considerations must be taken into account. Box 5 presents the prompt used in this step.

The task's context must include details about the characteristics of the data and the individuals providing it while omitting any unnecessary information that could introduce bias. Therefore, aspects such as hypotheses or research questions should preferably be excluded or described only in broad terms. Regarding the role, the coder can be described as an assistant to a team of social scientists, collaborating on the classification of unstructured data for analysis. For the task description, the prompt should explain that the coder must determine whether each item belongs to a given category. Additionally, researchers may specify rules or requirements coders must follow when classifying items. For example, they can clarify whether items should be assigned to only one or multiple categories, what type of information should be prioritized for classification, or how the classification procedure should be conducted. Finally, the expected format and coding scores must be specified. In this case, the best approach is for coders to present their responses in a tabular format, with the score for each category in a separate column.

For LLM, it is recommended to run the classification five times and assign a score to a category only if it receives the same score at least three times. It demonstrates robustness and confirmability and also prevents hallucinations (Smirnov, 2024).

Box 5. Prompt for classification.

---

**Context**

We are studying the pursuit and achievement of personal goals. Personal goals are defined as the outcomes a person aims to achieve, guided by their needs, interests, values, and aspirations.

We conducted a longitudinal study in which participants described their personal goals, and weeks later, we evaluated their progress. To analyze the kinds of goals people pursue, we need to classify the reported goals into predefined categories.

Each participant provided:

- A basic description of their goal (what they want to achieve).
- Additional information (why the goal is important, how long they have been pursuing it, and how they feel about it).

You will see a table containing information about personal goals from a group of individuals. The "goal" column presents the basic description of a goal along with any additional information they provided about its relevance.

**Role**

You are an AI assistant helping a social researcher analyze personal goals.

---

**Task Definition**

Your task is to classify and score each personal goal based on the following categories:

1. **Education and Learning:** Goals related to completing educational programs, acquiring new knowledge, abilities, or learning new skills (e.g., Finish my university studies because it is the career I chose and have always liked; Learn English to communicate my products to my audience and for job opportunities; learn how to drive).
2. **Financial Stability and Independence:** Goals focused on achieving financial security, independence, or improving economic conditions (e.g., Achieve financial stability to live more freely without worrying about money; Get out of debt to improve my economy and gain financial freedom).
3. **Career and Professional Development:** Goals related goals related to professional and work-life development. This would include job promotions, career changes, transitioning out of unemployment, and starting independent work through personal businesses or entrepreneurial ventures (e.g., Change jobs to experience new challenges and grow professionally;start my own business to provide more peace and well-being to my family).
4. **Health and Well-being:** Goals aimed at improving physical health, mental health, or overall well-being (e.g., Lose weight for health reasons and to feel strong for myself and my daughters; Exercise more to improve health and physical appearance).
5. **Family and Relationships:** Goals centered around improving family life, relationships, or supporting family members (e.g., Improve relationship with my son to maintain harmony and trust; Spend more quality time with my partner).
6. **Travel and Exploration:** Goals involving traveling to new places for leisure, cultural experiences, or personal fulfillment (e.g., Travel to Japan to learn about a new culture and see the world beyond my area; Travel around the world to achieve a higher status in my professional life).
7. **Material Acquisition:** Goals aimed at acquiring, retaining, or maintaining material goods. These may include purchasing a first or second home, vehicles, household appliances, leisure or personal enjoyment items, as well as home repairs or improvements (e.g.,Buy my own house; Buy a new car; Finish purchasing the furniture for my apartment.).
8. **Personal Development and Fulfillment:** Goals aimed at personal growth, self-improvement, or achieving a sense of fulfillment (e.g., To be a better person every day to feel happy and improve daily; Stop procrastinating to engage in activities that help me grow personally and professionally).
9. **Not Applicable:** Responses that are meaningless, nonsensical, or do not fit into any relevant category (e.g., Develop telepathy through meditation and practice; None, I don't have to give opinions about my goals).
10. **Orphans:** All goals that, although they make sense and are perfectly intelligible, do not fit into any of the categories.

Classification Rules

Each goal description can contain one or multiple goals. Classify the goals as follows:

1. Main Category (Score = 2).
    - The ultimate goal the person wants to achieve.
    - Each goal description must have one and only one main category.

> 2. Intermediate Category (Score = 1).
>    - Any additional goal that serves as a step toward achieving the main goal.
>    - A goal description may have one, multiple, or no intermediate categories.
> 3. Irrelevant Categories (Score = 0).
>    - Categories that do not apply to the goal description.
>    - Any category not assigned a 1 or 2 should be scored 0.
>
> Example:
> For the goal "saving to buy a house", assign:
>
> - 2 in the Material Acquisition category (ultimate goal: buying a house).
> - 1 in the Financial Stability and Independence category (saving is a step toward buying a house).
> - 0 in all other categories.
>
> Whenever possible, classify the goal based only on the basic description.
> Use additional information only when the description is unclear, ambiguous, or unintelligible.
>
> **Expected Output**
>
> - Present the responses in a tabular format.
> - Each goal description should have a score (0, 1, or 2) for each category in a separate column.

Researchers must evaluate two key aspects: (1) intercoder reliability, to ensure that the taxonomy is clear enough to prevent misunderstandings during classification, and (2) comprehensiveness, to assess whether the categories adequately capture all items in the dataset.

To assess the first aspect, a common strategy for testing intercoder reliability is to calculate coders' percentage of agreement. However, coders may agree on the classification of an item due to chance rather than actual agreement. Therefore, various reliability indices have been developed to account for chance. Examples include Cohen's Kappa, Fleiss' Kappa, the Intraclass Correlation Coefficient (ICC), and Krippendorff's Alpha (see in the Supplemental material a R script for calculating intercoder reliability with all those indices and a brief paper about the usage and interpretation of each index).

Reliability issues may be specific to certain categories rather than affecting the entire taxonomy. Therefore, researchers may find it useful to calculate reliability for each category separately to identify those requiring adjustments. The supplemental material includes an R script for preparing the data frame and computing several indices both across all categories and for individual categories.

To assess comprehensiveness, we recommend including an additional category in the taxonomy called "Orphans", which allows classifiers to capture items that do not fit in any other category. After classification, researchers should review the items assigned to the Orphans category. If a substantial number of items share a clear theme relevant to research purposes, the taxonomy can be modified by introducing a new category to encompass them. Conversely, if fewer than 5% of the items fall into the Orphans category, researchers can reasonably conclude that their taxonomy is sufficiently comprehensive.

To test our taxonomy, we selected a subset of our goal dataset. Given that our taxonomy includes 10 categories, we sampled 150 personal goals, which resulted in a total of 1,500 observations for goal-category evaluations. Using the prompt detailed in Box 5, we asked GPT and two research assistants to classify these goals. As mentioned previously, we ran the classification five times and established the final category for each item by verifying whether it was classified at least three times into the same category. Regarding reliability, we found a good agreement among the three coders (ICC = .824). In terms of comprehensiveness, we did not identify any orphan items, so we did not have to include anything additional.

**Step 7. Make Final Adjustments (if Necessary)**

If intercoder reliability metrics indicate a suboptimal level of agreement, researchers must review discrepancies among coders and refine the taxonomy until they reach an acceptable level of consistency. However, researchers usually balance reliability with the time and resources available to determine what degree of agreement is sufficient. At this stage, coders should discuss their disagreements and identify the reasons behind their divergences in categorization. Inter-coder discussions should reveal the main sources of disagreement. Usual causes are (1) ambiguous category descriptions and (2) lack of clarity about the boundaries of categories. Depending on the source of disagreement, researchers can do one or more of the following:

1. *Revise category descriptions* to make them clearer and more precise. This can be done by adjusting category labels, rephrasing definitions, or adding illustrative cases or examples.
2. *Add classification rules* to make the taxonomy clearer when specific types of cases should fall into one category rather than another, or how certain categories should relate.

Metrics per category are particularly useful at this stage, as they provide insights into necessary adjustments. A low or moderate overall intercoder reliability score may result from a specific category with a high degree of disagreement, suggesting that refinements should focus on that category rather than the entire taxonomy.

Researchers can also consider adding new categories if multiple items with similar content have been classified as orphans. However, it is crucial to assess whether these new categories are relevant enough to be incorporated into the taxonomy. If numerous items are categorized as orphans but grouping them into a specific category does not meaningfully contribute to the research objectives, researchers may opt to retain them in a "trash" category that encompasses all irrelevant items.

In our example, even though we obtained a good agreement index in our initial test, research assistants involved in the categorization expressed concerns regarding certain categories and ambiguities. In response, we discussed some disagreements and made several adjustments to the taxonomy based on their recommendations. Most adjustments involved introducing classification rules specifying when to assign a goal to a category, what score to give, and when to exclude it. Other rules addressed whether to rely solely on the basic description or include additional context, and how to handle cases where participants listed multiple but unrelated goals. After refining the taxonomy (Box 6), we conducted a new classification using a different data subset and obtained a slightly higher reliability index (ICC = .827).

Box 6. Refined version of taxonomy

**Context**

We are studying the pursuit and achievement of personal goals. Personal goals are defined as the outcomes a person aims to achieve, guided by their needs, interests, values, and aspirations.

We conducted a longitudinal study in which participants described their personal goals, and weeks later, we evaluated their progress. To analyze the kinds of goals people pursue, we need to classify the reported goals into predefined categories.

Each participant provided:

- A basic description of their goal (what they want to achieve).
- Additional information (why the goal is important, how long they have been pursuing it, and how they feel about it).

You will see a table containing information about personal goals from a group of individuals. The "goal" column presents the basic description of a goal along with any additional information they provided about its relevance.

**Role**

You are a research assistant helping a social researcher analyze personal goals.

**Task Definition**

Your task is to classify and score each personal goal based on the following categories:

1. **Education and Learning:** Goals related to completing educational programs, acquiring new knowledge, abilities, or learning new skills (e.g., Finish my university studies because it is the career I chose and have always liked; Learn English to communicate my products to my audience and for job opportunities; learn how to drive).
2. **Financial Stability and Independence:** Goals focused on achieving financial security, independence, or improving economic conditions (e.g., Achieve financial stability to live more freely without worrying about money; Get out of debt to improve my economy and gain financial freedom).
3. **Career and Professional Development:** Goals related goals related to professional and work-life development. This would include job promotions, career changes, transitioning out of unemployment, and starting independent work through personal businesses or entrepreneurial ventures (e.g., Change jobs to experience new challenges and grow professionally; start my own business to provide more peace and well-being to my family).
4. **Health and Well-being:** Goals aimed at improving physical health, mental health, or overall well-being (e.g., Lose weight for health reasons and to feel strong for myself and my daughters; Exercise more to improve health and physical appearance).
5. **Family and Relationships:** Goals centered around improving family life, relationships, or supporting family members (e.g., Improve relationship with my son to maintain harmony and trust; Spend more quality time with my partner).

6. **Travel and Exploration:** Goals involving traveling to new places for leisure or cultural experiences (e.g., Travel to Japan to learn about a new culture and see the world beyond my area; Travel around the world to achieve a higher status in my professional life).
7. **Material Acquisition:** Goals aimed at acquiring, retaining, or maintaining material goods. These may include purchasing a first or second home, vehicles, household appliances, leisure or personal enjoyment items, as well as home repairs or improvements (e.g., Buy my own house; Buy a new car; Finish purchasing the furniture for my apartment.).
8. **Personal Development and Fulfillment:** Goals aimed at personal growth, self-improvement, spiritual development or strengthen life skills to be a better person (e.g., To be a better person every day to feel happy and improve daily; Stop procrastinating to engage in activities that help me grow personally and professionally).
9. **Not Applicable:** Responses that are meaningless, nonsensical, or do not fit into any relevant category (e.g., Develop telepathy through meditation and practice; None, I don't have to give opinions about my goals).
10. **Orphans:** All goals that, although they make sense and are perfectly intelligible, do not fit into any of the categories.

**Expected Output**

Present the responses in a tabular format.
Each goal description should have a score (0, 1, or 2) for each category in a separate column.

Each goal description can contain one or multiple goals. Classify the goals as follows:

1. Main Category (Score = 2).
    - The ultimate goal the person wants to achieve.
    - Each goal description must have one and only one main category.
2. Intermediate Category (Score = 1).
    - Any additional goal that serves as a step toward achieving the main goal.
    - A goal description may have one, multiple, or no intermediate categories.
3. Irrelevant Categories (Score = 0).
    - Categories that do not apply to the goal description.
    - Any category not assigned a 1 or 2 should be scored 0.

Example:
For the goal "saving to buy a house", assign:

- 2 in the Material Acquisition category (ultimate goal: buying a house).
- 1 in the Financial Stability and Independence category (saving is a step toward buying a house).
- 0 in all other categories.

**Classification rules**

1. **Each category can only receive one score per goal (either 0, 1, or 2).**
2. **Basic description vs. additional information:** Always consider both the **basic description** and the **additional information** when categorizing a goal. However, if the **additional information** is **contradictory** to the basic description (e.g., *"My goal is to travel to Spain"* vs. *"because I deeply love my home country, Colombia"*), or **unrelated** to it (e.g., *"Complete my master's degree* *with excellent performance"* vs. *"My family's health is my top priority"*), use **only the basic description** to categorize the goal.
3. **Multiple, unrelated or non-subordinated objectives:** If a goal includes multiple, unrelated or non-subordinated objectives (e.g., *"I want to get a better job and lose weight"*), consider both the basic description and the additional information to determine which one is the main ultimate goal (e.g., *"Basic description: I want to get a better job and lose weight. Additional information: I want a better that challenges me"*) and:
    - Assign a score of **2** to the main ultimate goal.
    - Assign a score of **1** to any intermediate goals that clearly support the main goal.
    - Assign a score of **0** to any goal that is unrelated or not clearly subordinated to the main one.
    In the example above, score **Career and Professional Development = 2**, and **Health and Well-being = 0**.
4. **Two ultimate ends:** If a goal seems to have two distinct ultimate ends (e.g., *Basic description: "Travel to the United States." Additional information: "I want to travel to find a good job and spend time with my family"*):
    - Assign a **2** to the ultimate end mentioned first.
    - Assign a **1** to the second ultimate end.

   In the example above, score **Career and Professional Development = 2**, and **Travel and Exploration** and **Family and Relationships = 1**.

5. **Material acquisition vs. financial goals:** Acquiring an object or service does not imply Financial Stability and Independence unless the person explicitly mentions saving money or taking a loan to make the purchase. If the financial means are mentioned (e.g., *"save money to buy a car", "get a loan for a house"*), assign **1** to **Financial Stability and Independence** and 2 to **Material Acquisition**. If only the acquisition is mentioned, without any reference to financial means, assign a score of 2 to **Material Acquisition** only.

6. **Entrepreneurship and independence:** If someone aims to achieve professional independence through entrepreneurship, classify the goal under **Career and Professional Development** (e.g., "I want to become independent by opening my own business"). If the person refers to gaining financial independence from others (e.g., parents, partner, financial institutions), classify it under **Financial Stability and Independence** (e.g., "I want to stop depending financially on my parents").

7. **Skill development and domain specificity:** When a goal involves acquiring, developing, or improving skills, classify it based on the **type of skill.** If the skill is clearly tied to a domain (e.g., academic, financial, occupational, interpersonal), assign it to that domain. If the skill is not domain-specific, classify it under **Personal Development and Fulfillment**. For example:

    - *"Improve my financial skills"* → **Financial Stability and Independence**.

    - *"Improve my communication skills" (with no domain specified)* → **Personal Development and Fulfillment**.

8. **Personal growth and self-improvement:** Do **not** classify a goal under **Personal Development and Fulfillment** just because it mentions "achievement", "satisfaction", or "growth". Only use this category when the goal explicitly focuses on **personal or spiritual**

> **growth**, or on becoming a better person (e.g., *"be the best version of myself"*, *"develop more empathy"*). If growth refers to another domain (e.g., *"complete my master's degree"*, *"improve my finances"*, *"communicate better with my family"*), assign it to the relevant domain instead. For example:
>
> *"Complete my master's degrees"* → **Education and Learning**.
> *"improve my finances"* → **Financial Stability and Independence**
> *"communicate better with my family"* → **Family and Relationships**
> *"be the best version of myself"* → **Personal Development and Fulfillment**.
>
> 9. **Habits and organization:** If the goal involves general improvement in personal habits or organization (e.g., *"stop procrastinating"*, *"organize my routines better"*), classify it under **Personal Development and Fulfillment**. If the habits or organization are clearly linked to a specific domain (e.g., study, work, exercise), classify the goal in that domain only. For example:
>
> *"stop procrastinating"* → **Personal Development and Fulfillment**.
> *"stop procrastinating on my thesis writing"* → **Education and Learning**.

**Step 8. Apply the Taxonomy**

Once the taxonomy has been satisfactorily tested and coders have reached the expected level of agreement, the taxonomy may be used to classify the entire dataset. The output of this final step would be a dataframe containing all observations, along with scores denoting whether each observation belongs to each taxonomy category. With this information, researchers can calculate the frequency of observations per category and explore the relationship between category membership and other variables. In our case, the most common personal goals were Carrer and professional development (21.4%), Education and learning (18.9%), and Material acquisitions (18.9%).

This final step, in turn, should lead to a discussion of the taxonomy's theoretical implications and its broader relevance. What has been learned? How does this taxonomy compare to those proposed in other studies? What are the key differences? How do the results of this bottom-up process differ from those produced through a top-down approach? When comparing our example with the most well-known goal taxonomies, we observed several key differences, despite certain similarities. Specifically, our taxonomy does not include some categories found in a prior taxonomy (Grouzet et al.,2005), such as popularity and hedonism, but introduces other categories such as education and learning, or career and professional development. These differences are likely explained by the theoretical framework underpinning Grouzet's taxonomy, which is heavily focused on the dimensions of intrinsic vs. extrinsic goals and physical self vs. self-transcendence—two dimensions that, although important, were not the main focus of our project.

**Summary and discussion**

We presented an eight-step procedure for developing, testing, and applying a taxonomy to analyze qualitative data using an LLM. The first step is designing a prompt to generate an initial taxonomy aligned with research objectives and predefined requirements. The second step consists of executing a Python script to send the query to GPT, which results in an initial taxonomy. The third step is evaluating this output against taxonomy quality criteria, identifying limitations such as redundant or narrowly scoped categories. To refine the taxonomy based on those shortcomings and obtain a good enough version for testing, the fourth step asks researchers to revise the initial prompt (if appropriate), and the fifth step asks them to adjust the taxonomy itself. For testing, in step 6 two classifiers (human or LLM-based) independently categorize a subset of data using the refined taxonomy to assess its clarity (via intercoder reliability) and coverage. If inter-rater reliability is not sufficiently high, in step 7, researchers adjust the taxonomy to obtain satisfactory intercoder agreement indices and broad comprehensiveness. The final step is applying the refined taxonomy to the full dataset.

This exercise illustrates how this tutorial can be a valuable tool to guide social scientists in generating, testing, and applying a taxonomy to analyze unstructured data using LLMs to save time and reduce effort while simultaneously mitigating potential biases. Unlike a fully automated process, analyzing unstructured data with LLMs requires iterative collaboration between the LLM and the human researcher to ensure quality and reliability. Nevertheless, despite the need for human involvement, incorporating an LLM significantly streamlines the process compared to manual methods, putting human researchers in the role of managers of highly-capable assistants, and allowing the former to choose which tasks (taxonomy creation and evaluation, and data classification) to leave in the hands of artificial systems. Previously, the extensive time and effort required for the qualitative analysis of large volumes of unstructured data made this process either prohibitively expensive or extremely time-consuming, thereby limiting their use. Now, this is no longer a limitation. Social science researchers can use LLMs as assistants to streamline their qualitative analyses without compromising quality. Using tools like this tutorial, researchers are now able to leverage such tools to analyze vast amounts of information that were previously unmanageable.

Applications of LLMs in qualitative analysis should still be approached with caution. Models such as GPT, Claude, and DeepSeek are relatively new and continuously evolving. We are only beginning to explore their potential and limitations. For example, several studies have examined LLM hallucinations: fabricated outputs that, although seemingly plausible, diverge from the given input, its context, or conventional knowledge (Liu et al., 2024; Zhang et al., 2023; Huang et al., 2025). Even though each step in this tutorial was designed to mitigate the risk of hallucinations (by including clear, context-rich prompts, simple inputs, thorough output evaluation, and multiple classification rounds), the possibility for hallucination remains. In our use of these steps for taxonomy design, we have yet to come across the first such case. The use of reliability-enhancing tools like this tutorial is crucial for the application of LLMs to qualitative analysis is highly concerning since hallucinations and similar issues could completely alter research findings and conclusions.

Another crucial ethical consideration regarding the use of LLMs in qualitative analyses is the possibility of biased responses. LLMs and other machine learning systems are known to produce biased outputs related to social categories such as gender, ethnicity, race, religion, and political orientation (Angwin et al., 2016; Fazelpour & Danks, 2021; Kasirzadeh, 2022; Fulgu & Capraro, 2024; Zack et al., 2024; Motoki et al., 2023; Liang et al., 2021; Liu et al., 2022). These biases raise concerns about how training data might reflect and amplify human prejudices, as well as how fine-tuning—shaped by developers' choices—could introduce additional biases. Human judges may be equally or even more biased, though, so further research is needed to better understand how to reduce overall bias in human-LLM collaborative classification. That said, familiar measures can be adopted to minimize the risk of biased outputs, such as anonymizing the data and avoiding the inclusion of sociodemographic information when it is irrelevant to the classification task and importantly ask participants for their consent to analyse their responses with AI tools.

Some could worry about the possibility that LLMs will ultimately replace the human workforce. To tackle this concern, we designed this guide with human empowerment in mind: it allows researchers to maintain meaningful human control (Santoni di Sio & Van den Hoven, 2018) over the entire process. Following these steps, researchers directly control their tools, assess when the tools are functioning properly or misfiring, and choose what aspects of the tasks to delegate and which should be done by humans. Additionally, the main objective is to make it possible for teams of human researchers to rigorously analyze datasets that were previously unfeasible due to resource constraints.

**Conclusions**

In this tutorial, we have presented a step-by-step guide to developing, testing, and applying a taxonomy to analyze unstructured data through an iterative process. This process integrates the collaborative work of researchers and LLMs, where the former define the research goals, configure a series of prompts for the LLMs, evaluate outputs, and make decisions at each stage, while the latter perform the most time- and effort-intensive tasks, such as reviewing and integrating large volumes of unstructured data or classifying a large number of items or text units into categories. By analyzing two datasets for different research purposes, we have demonstrated that this joint work can make qualitative analyses more efficient without compromising quality. Although these results are promising for social research, we caution against an uncritical adoption of LLMs in the scientific process, as they are not perfect, and we are only beginning to understand their scope, limitations and risks.

**Funding**

This project was funded by the John Templeton Foundation grant 62622, *Self-Control in Context*, and by Teaching and Researching Equitable Economics from the South (TREES).

**Acknowledgments**

We would like to thank Mike Angstadt and Aman Taxali for their help in drafting the Python code and for kindly answering questions about using the GPT 4.0 API. We are also grateful to Juana Ospitia, Isabella González Viatella and Marian Dorado Toro for their valuable assistance with classification.

Hruschka, D. J., Schwartz, D., St. John, D. C., Picone-Decaro, E., Jenkins, R. A., & Carey, J. W. (2004). Reliability in coding open-ended data: Lessons learned from HIV behavioral research. *Field Methods, 16*(3), 307–331. https://doi.org/10.1177/1525822X04266540

Huang, L., Yu, W., Ma, W., Zhong, W., Feng, Z., Wang, H., ... & Liu, T. (2025). A survey on hallucination in large language models: Principles, taxonomy, challenges, and open questions. *ACM Transactions on Information Systems, 43*(2), 1–55. https://doi.org/10.1145/3703155

Iliev, R., Dehghani, M., & Sagi, E. (2015). Automated text analysis in psychology: Methods, applications, and future developments. *Language and Cognition, 7*(2), 265–290. https://doi.org/10.1017/langcog.2014.30

Jeon, J., & Lee, S. (2023). Large language models in education: A focus on the complementary relationship between human teachers and ChatGPT. *Education and Information Technologies, 28*(12), 15873–15892. https://doi.org/10.1007/s10639-023-11834-1

Kasirzadeh, A. (2022). Algorithmic fairness and structural injustice: Insights from feminist political philosophy. *Proceedings of the 2022 AAAI/ACM Conference on AI, Ethics, and Society*, 349–356. https://doi.org/10.1145/3514094.3534188

Kasser, T., & Ryan, R. M. (1993). A dark side of the American dream: Correlates of financial success as a central life aspiration. *Journal of Personality and Social Psychology, 65*(2), 410–422. https://doi.org/10.1037/0022-3514.65.2.410

Kasser, T., & Ryan, R. M. (1996). Further examining the American dream: Differential correlates of intrinsic and extrinsic goals. *Personality and Social Psychology Bulletin, 22*(3), 280–287. https://doi.org/10.1177/0146167296223006

Kasser, T., & Ryan, R. M. (2001). Be careful what you wish for: Optimal functioning and the relative attainment of intrinsic and extrinsic goals. En P. Schmuck & K. M. Sheldon (Eds.), *Life goals and well-being: Towards a positive psychology of human striving* (pp. 116–131). Hogrefe & Huber.

Koo, T. K., & Li, M. Y. (2016). A guideline of selecting and reporting intraclass correlation coefficients for reliability research. *Journal of Chiropractic Medicine, 15*(2), 155–163. https://doi.org/10.1016/j.jcm.2016.02.012

Kunst, J. R., & Bierwiaczonek, K. (2023). Utilizing AI questionnaire translations in cross-cultural and intercultural research: Insights and recommendations. *International Journal of Intercultural Relations, 97*, 101888. https://doi.org/10.1016/j.ijintrel.2023.101888

Lee, P., Fyffe, S., Son, M., Jia, Z., & Yao, Z. (2023). A paradigm shift from "human writing" to "machine generation" in personality test development: An application of state-of-the-art natural language processing. *Journal of Business and Psychology, 38*(1), 163–190. https://doi.org/10.1007/s10869-022-09864-6
28

Sim, J., & Wright, C. C. (2005). The kappa statistic in reliability studies: Use, interpretation, and sample size requirements. *Physical Therapy, 85*(3), 257–268. https://doi.org/10.1093/ptj/85.3.257

Skiba, P. M., & Tobacman, J. (2008). *Payday loans, uncertainty, and discounting: Explaining patterns of borrowing, repayment, and default* (Working Paper No. 08–33). Vanderbilt University Law School. http://ssrn.com/abstract=1319751

Smirnov, E. (2024). Enhancing qualitative research in psychology with large language models: A methodological exploration and examples of simulations. *Qualitative Research in Psychology, 21*(1), 1–31. https://doi.org/10.1080/14780887.2024.2428255

Surden, H. (2023). ChatGPT, AI large language models, and law. *Fordham Law Review, 92*(5), 1941–1972.

Thirunavukarasu, A. J., Ting, D. S. J., Elangovan, K., Gutierrez, L., Tan, T. F., & Ting, D. S. W. (2023). Large language models in medicine. *Nature Medicine, 29*(8), 1930–1940. https://doi.org/10.1038/s41591-023-02448-8

World Health Organization. (2011). *Systematic review of the link between tobacco and poverty*. http://apps.who.int/iris/bitstream/10665/136001/1/9789241507820_eng.pdf

Zack, T., Lehman, E., Suzgun, M., Rodriguez, J. A., Celi, L. A., Gichoya, J., ... & Alsentzer, E. (2024). Assessing the potential of GPT-4 to perpetuate racial and gender biases in health care: A model evaluation study. *The Lancet Digital Health, 6*(1), e12–e22. https://doi.org/10.1016/S2589-7500(23)00225-X

Zhang, Z., Wang, Y., Wang, C., Chen, J., & Zheng, Z. (2024). LLM hallucinations in practical code generation: Phenomena, mechanism, and mitigation. *arXiv*. https://arxiv.org/abs/2409.20550
31